\documentclass[conference]{IEEEtran}
\IEEEoverridecommandlockouts
\usepackage[utf8]{inputenc} % allow utf-8 input
\usepackage[T1]{fontenc}    % use 8-bit T1 fonts
\usepackage{hyperref}       % hyperlinks
\usepackage{url}            % simple URL typesetting
\usepackage{booktabs}       % professional-quality tables
\usepackage{amsfonts}       % blackboard math symbols
\usepackage{nicefrac}       % compact symbols for 1/2, etc.
\usepackage{microtype}      % microtypography
\usepackage{xcolor}         % colors
\usepackage{graphicx}
\usepackage{amssymb}
\usepackage{booktabs}
\usepackage{colortbl}
\usepackage{multirow}
\usepackage{amsmath}

\definecolor{aliceblue}{rgb}{0.94, 0.97, 1.0}
\definecolor{lightgray}{rgb}{0.95, 0.95, 0.95}
\usepackage{cite}
\usepackage{amsmath,amssymb,amsfonts}
\usepackage{algorithmic}
\usepackage{graphicx}
\usepackage{textcomp}
\usepackage{xcolor}
\def\BibTeX{{\rm B\kern-.05em{\sc i\kern-.025em b}\kern-.08em
    T\kern-.1667em\lower.7ex\hbox{E}\kern-.125emX}}
\begin{document}

\makeatletter
\newcommand{\linebreakand}{%
      \end{@IEEEauthorhalign}
      \hfill\mbox{}\par
      \mbox{}\hfill\begin{@IEEEauthorhalign}
    }
\makeatother

\title{MambaMOT: State-Space Model as Motion Predictor for Multi-Object Tracking}

\author{\IEEEauthorblockN{Hsiang-Wei Huang, Cheng-Yen Yang, Wenhao Chai, Zhongyu Jiang, Jeng-Neng Hwang}
    \IEEEauthorblockA{\textit{Department of Electrical and Computer Engineering} \\
    \textit{University of Washington} \\
    \textit{Seattle, United States} \\
    \textit{\{hwhuang, cycyang, wchai, zyjiang, hwang\}@uw.edu}
}}

\maketitle

\begin{abstract}
In the field of multi-object tracking (MOT), traditional methods often rely on the Kalman filter for motion prediction, leveraging its strengths in linear motion scenarios. However, the inherent limitations of these methods become evident when confronted with complex, nonlinear motions and occlusions prevalent in dynamic environments like sports and dance. This paper explores the possibilities of replacing the Kalman filter with a learning-based motion model that effectively enhances tracking accuracy and adaptability beyond the constraints of Kalman filter-based tracker. In this paper, our proposed method \textbf{MambaMOT} and \textbf{MambaMOT\textsuperscript{+}}, demonstrate advanced performance on challenging MOT datasets such as DanceTrack and SportsMOT, showcasing their ability to handle intricate, non-linear motion patterns and frequent occlusions more effectively than traditional methods.

\end{abstract}
\section{Introduction}

Multi-object tracking is a fundamental task in computer vision, aims at tracking multiple target objects in a video stream, which has wide applications in video tasks~\cite{song2024moviechat,chai2023stablevideo,chai2023global,chai2024auroracap}. Among all the different trackers, motion-based trackers~\cite{zhang2022bytetrack,cao2023observation,bewley2016simple} are especially popular and can be easily applied to many real-world scenarios due to their simplicity and lower computational costs. Motion-based trackers~\cite{aharon2022bot,zhang2022bytetrack,cao2023observation,du2023strongsort,wojke2017simple,seidenschwarz2023simple,bewley2016simple} usually heavily rely on the prediction of Kalman filter~\cite{kalman1960new} during tracking, which demonstrates effectiveness when tracked objects have regular and linear movements, such as pedestrians~\cite{milan2016mot16,dendorfer2020mot20}. However, these trackers struggle to perform well on datasets~\cite{sun2022dancetrack,cui2023sportsmot} with objects having diverse and irregular motions. Several studies~\cite{cao2023observation,huang2024iterative} show that the linear motion assumption of the Kalman filter breaks down when tracking objects in dancing or sports scenarios. For this reason, we aim to explore stronger motion models that can provide more robust and adaptive motion predictions to improve the performance of Kalman filter-based trackers.

\begin{figure}[t]
\centering
\includegraphics[width=\linewidth]{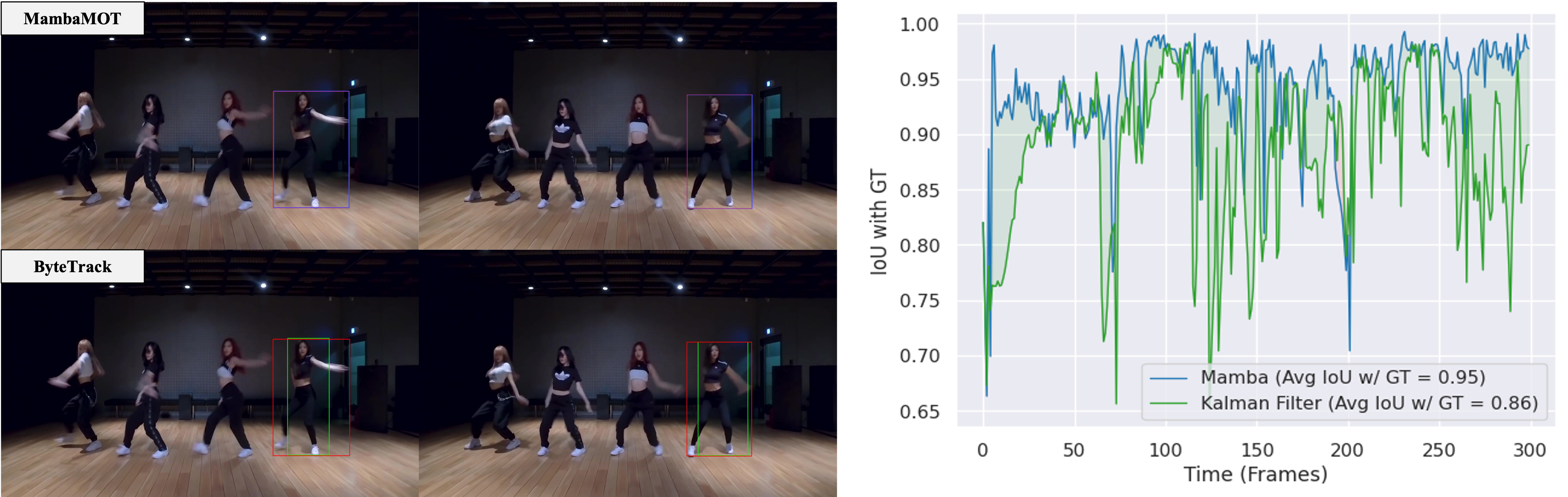}

\caption{A comparison between MambaMOT's and Kalman filter's tracking result. The left visualized results compare the predicted bounding boxes of our proposed MambaMOT and the state-of-the-art method ByteTrack, with the blue and green bounding boxes denoting the predicted bounding box of the target, and the red bounding box is the ground truth. The right figure compares the IoU between the prediction and ground truth bounding boxes; MambaMOT consistently demonstrates better location prediction accuracy compared to the Kalman filter. \textbf{Best viewed zoomed in and in color.}}
\label{fig:fig1}
\end{figure}

In this work, we seek to explore learning-based motion models as a replacement for the Kalman filter. Unlike the Kalman filter, which predicts an object's location using a linear motion assumption, the learning-based motion model adaptively predicts the tracklet's location in the next timestamp by leveraging tracklet information from previous frames. This enables the motion model to be more adaptive to each dataset's motion pattern, thus making more reliable predictions even when the target's motion is irregular. In this paper, we proposed MambaMOT, a pioneering online and real-time MOT approach that leverages exceptional context reasoning capabilities from the state-space model mamba~\cite{gu2023mamba} to conduct tracklet motion predictions, MambaMOT significantly improves tracking outcomes compared with existing motion-based tracker. Furthermore, we propose MambaMOT\textsuperscript{+}, which exploits the power of the state-space model to extract trajectory features and further boost the tracking performance. The proposed MambaMOT\textsuperscript{+} and MambaMOT achieve advanced performance across challenging multi-object tracking datasets such as DanceTrack~\cite{sun2022dancetrack} and SportsMOT~\cite{cui2023sportsmot}.
\section{Related Works}
\subsection{Tracking by Detection}
Most state-of-the-art trackers~\cite{aharon2022bot,zhang2022bytetrack,cao2023observation,seidenschwarz2023simple,du2023strongsort} adopt the tracking-by-detection paradigm. This approach involves applying an object detector to image frames, with the obtained detections then utilized for tracking. Many tracking-by-detection methods~\cite{bewley2016simple,wojke2017simple,zhang2022bytetrack,cao2023observation,kuan2024boosting,yang2024samurai} use the Kalman filter~\cite{kalman1960new} as a motion model to predict the location of tracklets in the next frame. The Intersection over Union (IoU) between the predicted boxes and detection boxes is used as the similarity metric for association. However, previous works~\cite{cao2023observation} have identified several limitations of the Kalman filter, including sensitivity to state noise and temporal error magnification. In scenarios of diverse and large motion, the Kalman filter's sensitivity to state noise can lead to a significant shift in final position estimation after just a few frames of being untracked. Thus, a fundamental solution to this problem is to incorporate a learning-based motion model that can adaptively predict an object's location based on the target's historical tracklet information.

\subsection{Motion Model for Multi-Object Tracking}
Most of the tracking by detection methods~\cite{bewley2016simple,wojke2017simple,zhang2022bytetrack,cao2023observation} incorporate motion model for association. However, several challenges including the tracking of objects with similar appearance and the motion modeling of objects with diverse motion~\cite{sun2022dancetrack,cui2023sportsmot} still remain unsolved. Even the recent benchmarks and challenges focusing on targets with diverse motion, most of the tracking by detection method~\cite{zhang2022bytetrack,aharon2022bot,zhang2021fairmot,wojke2017simple,bewley2016simple,du2023strongsort} still in favor of incorporating the Kalman filter~\cite{kalman1960new} as motion model. With the linear motion assumption of the Kalman filter, these methods inevitably fail on datasets like DanceTrack~\cite{sun2022dancetrack} and SportsMOT~\cite{cui2023sportsmot} with objects having diverse and irregular motion.
For this reason, we aim to explore different learning-based motion models and offer more robust alternatives to the tracking-by-detection paradigm under challenging tracking scenarios.

\subsection{Mamba}  

Building on the State-Space Model (SSM)~\cite{gu2021efficiently}, Mamba~\cite{gu2023mamba} is a novel, input-dependent, and hardware-efficient sequence model that outperforms transformers in both performance and efficiency. Mamba's design has inspired various applications, such as MambaEVT~\cite{wang2024mambaevt} and Mamba-FETrack~\cite{mambafetrack} for Visual Object Tracking (VOT) with event cameras, and MambaTrack~\cite{xiao2024mambatrack}, which uses a bi-directional architecture for motion modeling. In contrast, our MambaMOT harnesses Mamba’s efficient motion modeling specifically for Multi-Object Tracking (MOT). With its flexibility and computational efficiency, Mamba serves as a robust foundation for advancing object tracking across diverse modalities.

\begin{figure*}[!t]
\vspace{1em}
\centering
\includegraphics[width=0.8\linewidth]{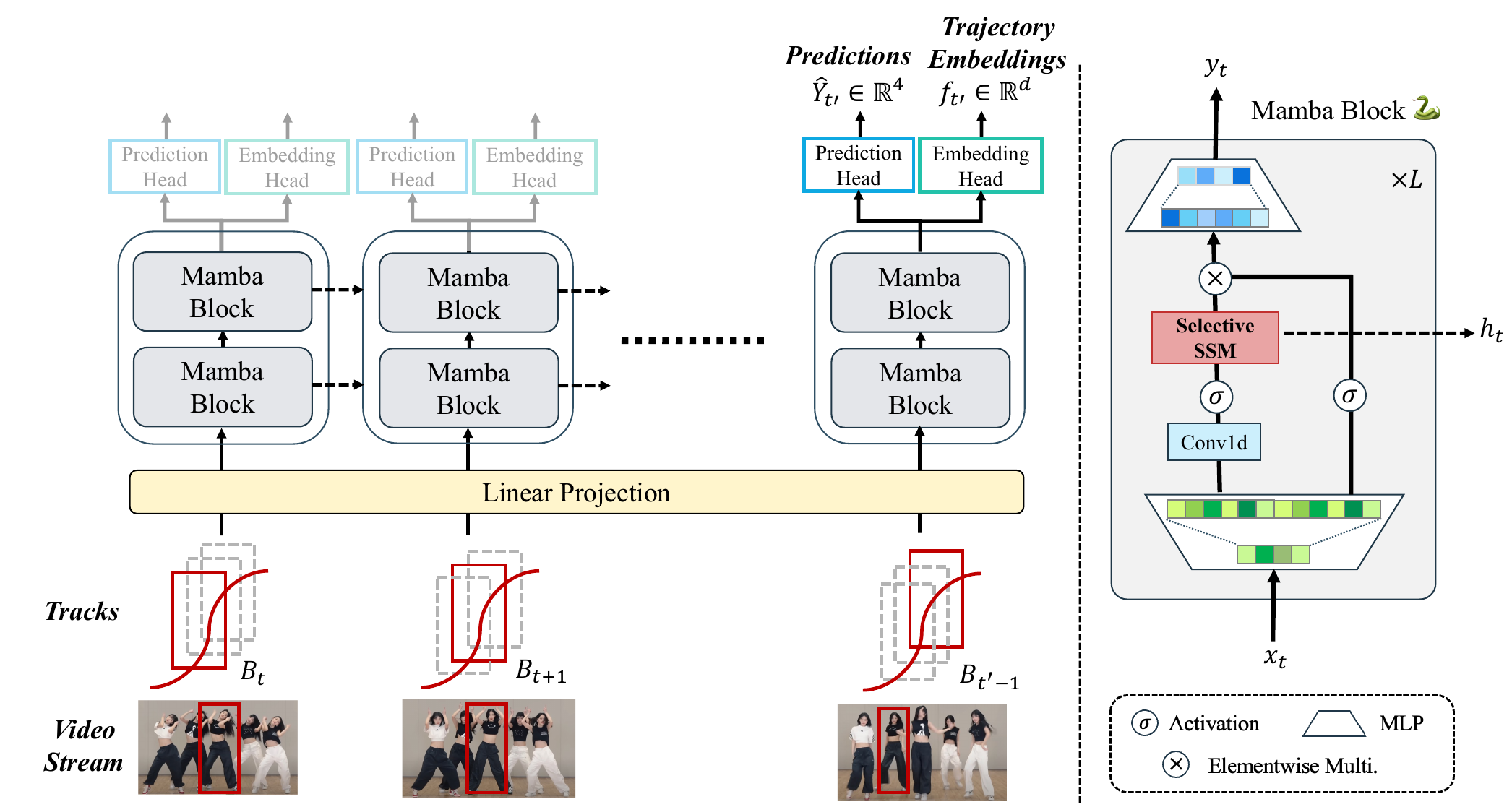}
\caption{\textbf{(Left)} 
The MambaMOT$^{+}$ architecture processes a sequence of bounding boxes from the same track through a linear projection layer for motion modeling. The model generates predictions and embeddings, updating the hidden state $h_T$ at each time frame. These predictions are used for detecting and matching tracks, while trajectory embeddings aid in merging tracklets. The detailed structure of the Mamba block is integral to this framework.}
\label{fig:arch}
\end{figure*}
\section{Method}
\subsection{Task Definition}
Given a tracklet's of the past $n$ frames $\{B_{t-n},...,B_{t-1}\} \in \mathbb{R}^{n \times [x,y,w,h]}$, the motion model aims at predicting the tracklet's location in the next time stamp ${B_{T}} \in \mathbb{R}^{1 \times [x,y,w,h]}$ based on the information from the past $n$ frames bounding boxes. The predicted bounding box will further be used for association with the detections $X_t$ from the next timestamp $t$.

\subsection{Motion Models}
% We explored the potential of learning-based motion model by implementing different architectures and comparing them with Kalman filter and other traditional tracking methods on several public multi-object tracking datasets \cite{milan2016mot16,sun2022dancetrack}.

We propose to implement the motion model using the original Mamba \cite{gu2023mamba} block, with an extra linear projection layer, which transforms the bounding box to a higher dimension and a local multi-layer perception for predicting the tracklet's next location based on a continuous system with parameters $A, B, C$:

\begin{gather}
\label{eq:yt}
    y(t) = C\cdot h(t)\\  
\label{eq:ht}    
    h(t) = A \cdot h(t-1) + B\cdot x(t).
\end{gather}

The Selective SSM (state-space model) is essentially the discrete representation $\bar{A}$, $\bar{B}$ of the continuous system $A$, $B$ by applying ZOH (zero-order hold) discretization formulation with a timescale parameter $\Delta$, i.e.,

% \begin{align*}
% \bar{A_t} &= \exp(\Delta A)  \\
% &= \frac{1}{1 + \exp(\text{-Linear}(x_t))} = \sigma(\text{-Linear}(x_t)) = 1 - \sigma(\text{Linear}(x_t)) \\
% \tilde{B}_t &= (\tilde{A})^{-1}(\exp(\Delta A) - I) \cdot \Delta B = (\exp(\Delta A) - I) = 1 - \tilde{A} \\
%             &= \sigma(\text{Linear}(x_t)).
% \end{align*}

\begin{align}
\bar{A_t} &= \exp(\Delta A) = 1 - \sigma(\text{Linear}(x_t)) \\
\bar{B}_t = (\bar{A}&)^{-1}(\exp(\Delta A) - I) \cdot \Delta B =  \sigma(\text{Linear}(x_t)),
\end{align}

\noindent where the discrete output $y_t$ and hidden state $h_t$ from Eq. \ref{eq:yt} and \ref{eq:ht} as:

\begin{gather}
    y_t = C\cdot h_t \quad \text{and} \quad h_t = \bar{A_t}\cdot h_{t-1} + \bar{B_t}\cdot x_t,
\end{gather}

\noindent then, by incorporating selective State Space Models (SSMs) into a streamlined end-to-end neural network architecture that foregoes traditional attention mechanisms as described in the original work, we can significantly increase the inference speed while offering linear scalability with sequence length. Similar to the implementation of the most recurrent models, we compute and propagate the hidden states and finally produce the tracklet's next location in the final output by feeding the output to the prediction head as illustrated in Fig.~\ref{fig:arch}.

% $\textbf{MLP}_{pred}$:

% \begin{equation}
% \label{eq:pred_head}
%     \mathbf{Y}_{t} = \textbf{MLP}_{pred}(y_t).
% \end{equation}

% To our knowledge, this is the first method that incorporates a state-space model for motion modeling in an MOT task. We incorporate this motion model with the association method BYTE \cite{zhang2022bytetrack} and named this proposed architecture \textbf{MambaMOT}.

\subsection{Motion Prediction}

The prediction head aims to predict the tracklet's location in the next frame. We compute and propagate the hidden states and finally produce the tracklet's next location $\mathbf{Y}_{t}$ as final output by feeding the discrete output $y_t$ to the prediction head $\textbf{MLP}_{pred}$:

\begin{equation}
\label{eq:pred_head}
    \mathbf{Y}_{t} = \textbf{MLP}_{pred}(y_t).
\end{equation}

To enable accurate bounding box prediction, we apply Generalized IoU loss (GIoU)\cite{rezatofighi2019generalized} and MSE loss between the predicted bounding box and the ground-truth bounding box. By minimizing these losses, the model learns to generate bounding box predictions closer to the ground truth in terms of spatial coordinates. The final loss for $\textbf{MLP}_{pred}$ is: $\mathcal{L}_{\text{pred}} = \mathcal{L}_{\text{giou}} + \mathcal{L}_{\text{mse}}.$

% \begin{equation}
% \mathcal{L}_{\text{pred}} = \mathcal{L}_{\text{giou}} + \mathcal{L}_{\text{mse}}.
% \end{equation}

To show learning-based motion model can conduct better motion modeling than the Kalman filter, MambaMOT incorporates the data-association method BYTE proposed in ByteTrack  \cite{zhang2022bytetrack}. Unlike our MambaMOT, which uses Mamba as the motion model, ByteTrack utilized the Kalman filter to predict tracklet location in the next frame. The association is conducted between the predicted location and the detection using the Hungarian algorithm. A visualization in Fig.~\ref{fig:fig1} showcase the effectiveness of MambaMOT compared to Kalman filter-based tracker.

\subsection{MambaMOT\textsuperscript{+}: Unleashing the Power of Mamba}
\subsubsection{Motion Pattern in Hidden State.}
A strong property of the structured state-space model is its ability for long-context reasoning \cite{gu2023mamba,gu2020hippo,gu2021efficiently}, which has demonstrated superior performance over transformer model on datasets like long-range arena \cite{tay2020long}. 
% The reason for Mamba and other structured state-space models to achieve superior long-context reasoning is their ability to preserve previous input signals in the hidden state and propagate over time. This property enables Mamba to capture the motion patterns and predict the next location based on the object's historical trajectory, resulting in significant performance gains in online tracking scenarios. During the process of motion modeling, the hidden states of Mamba contain rich motion patterns of the tracklet. 
To further exploit the ability of Mamba to exploit the tracklet's motion pattern in the hidden state, we further proposed MambaMOT\textsuperscript{+}, which aims at unleashing the power of Mamba by extracting the trajectory's motion pattern as trajectory feature, and further connecting tracklets with similar trajectory features to boost the tracking performance.

\subsubsection{Trajectory Representation.}
MambaMOT\textsuperscript{+} incorporates an extra trajectory embedding head as shown in Fig.~\ref{fig:arch}, which takes the result from the mamba block $y_t$ as input into another trainable multilayer perceptron $\textbf{MLP}_{emb}$ to get the trajectory feature $\mathbf{f}_{t}$. The trajectory head captures global representations of the tracklet's trajectory information from the hidden states, and thus contains rich motion patterns and position information that can be further used to connect those fragment tracklets during the tracking process.

\begin{equation}
\label{eq:emb_head}
    \mathbf{f}_{t} = \textbf{MLP}_{emb}(y_t)
\end{equation}

The training objective of trajectory representations is to minimize the embedding features' cosine distance between the same tracklet while maximizing between different ones. We applied the cosine embedding loss $\mathcal{L}_{\text{cos}}$ to achieve this objective. A pair of trajectories from tracklet $i$ and tracklet $j$ will be sampled during each loss backward, and the cosine embedding loss $\mathcal{L}_{\text{cos}}$ will be calculated following:

\begin{equation}
\mathcal{L}_{\text{cos}}(i, j) = \begin{cases}
1 - \text{cos}(\mathbf{f}_i, \mathbf{f}_j), & \text{if i $=$ j}\\
\max(0, \text{cos}(\mathbf{f}_i, \mathbf{f}_j)), & \text{if i $\neq$ j}
\end{cases}
\end{equation}

\noindent where $\mathbf{f}_i$ represents the global head's predicted features by forwarding the trajectory sampled from tracklet $i$ through the mamba blocks. We jointly train the bounding box prediction head and global features head by adding the loss together. The end-to-end loss is thus defined as: $\mathcal{L}_{\text{total}} = \mathcal{L}_{\text{pred}} + \mathcal{L}_{\text{cos}}.$
% \begin{equation}
% \mathcal{L}_{\text{total}} = \mathcal{L}_{\text{pred}} + \mathcal{L}_{\text{cos}}.
% \end{equation}

\subsubsection{Tracklet Merging.}
Many previous works used a deep learning-based model~\cite{du2023strongsort,huang2023observation,yang2024sea,sun2024gta,yang2024online} to conduct tracklet merging based on the appearance or motion pattern similarity of tracklet pair. A prevalent approach among these works involves employing Siamese networks for this task. While these tracklet merging networks can boost the tracking performance by connecting the fragment trajectories from the same identity, the stand-alone Siamese network needs to compute the similarity between each tracklet pair and further introduces extra $O(N^2)$ computational cost, with $N$ denoting the number of tracklets. Different from these works, our proposed MambaMOT\textsuperscript{+} extracts the trajectory feature of tracklet using the same model for motion prediction during tracking process, which prevents the need for training another separate tracklet merging network, furthermore, MambaMOT\textsuperscript{+} operates its forward pass among each tracklet instead of tracklet pair, which reduec the computaional cost from $O(N^2)$ to $O(N)$ compared with Siamese-based methods.
After the tracking is finished, the trajectory feature of each tracklet will be compared using cosine similarity, we further use hierarchical clustering to merge the tracklet with similar trajectory features and boost the tracking performance.
\section{Experiments}

% \subsection{Datasets}
% We conducted experiments using the DanceTrack dataset~\cite{sun2022dancetrack} and the SportsMOT dataset~\cite{cui2023sportsmot}, which poses significant challenges due to the presence of targets with similar appearances and diverse motion patterns. These datasets offer a more complex environment, providing a robust benchmark for evaluating the performance of motion models.

\begin{figure}[t]
\centering
\includegraphics[width=\linewidth]{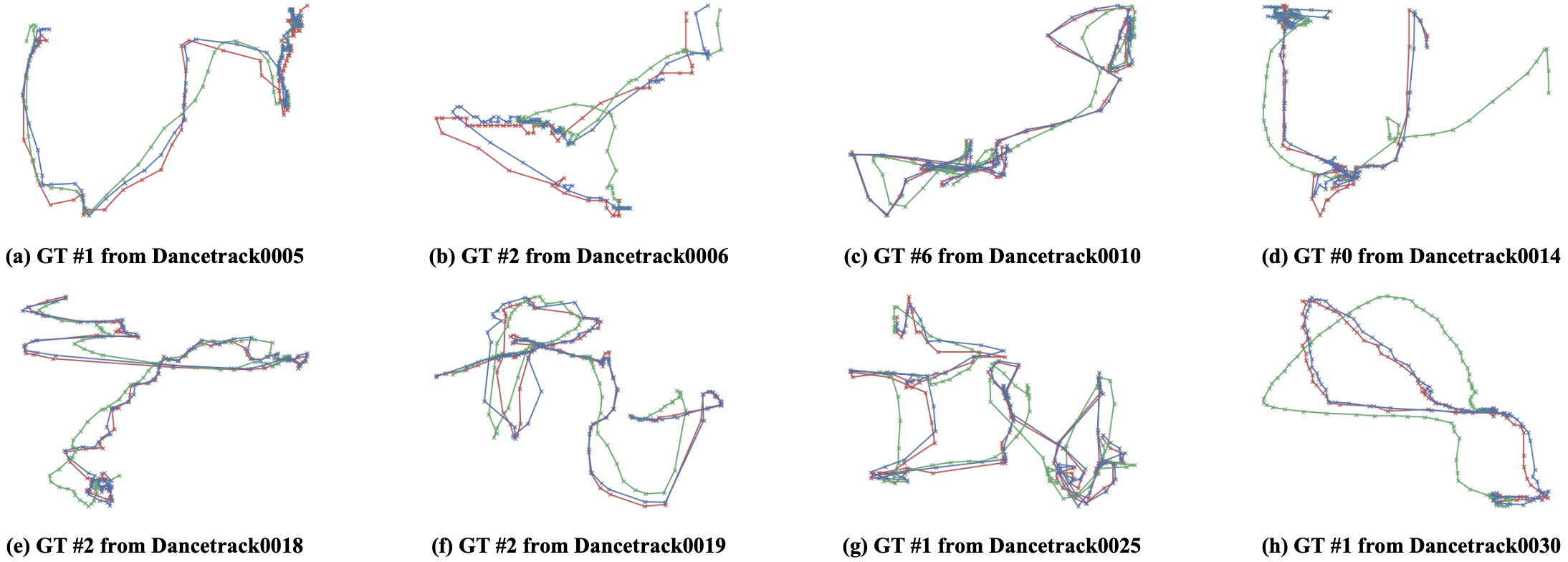}
\caption{Some randomly sampled trajectory visualizations between MambaMOT (Blue) and ByteTrack's (Green) compared with Ground truth trajectory (Red) on DanceTrack dataset. MambaMOT demonstrate better tracking accuracy in most of the cases.}
\label{fig:fig3}
\end{figure}

\subsection{Implementation Details}

\noindent{\textbf{Datasets.}} We conducted experiments using the DanceTrack dataset~\cite{sun2022dancetrack} and the SportsMOT dataset~\cite{cui2023sportsmot}, which poses significant challenges due to the presence of targets with similar appearances and diverse motion patterns. These datasets offer a more complex environment, providing a robust benchmark for evaluating the performance of motion models.

\noindent{\textbf{Detector.}} For fair comparison on DanceTrack datasets, we use the same detections obtained from ByteTrack's YOLOX model\cite{ge2021yolox}. For the SportsMOT~\cite{cui2023sportsmot} dataset, we train a YOLOX detector on the SportsMOT training set following the same training recipe of ByteTrack~\cite{zhang2022bytetrack}.

\vspace{2pt}

\noindent{\textbf{Trajectory Data.}} The training data  is collected from the trajectories in different multi-object tracking datasets, including MOT17, DanceTrack and SportsMOT. In the training process, the trajectories are randomly selected with the length between $(2,n)$, where $n$ is the max input length. We use padding to fill the tracklets length to $n$ when the sampled tracklets have a length smaller than $n$. The next bounding box location of these sampled tracklets is the ground-truth bounding box.

\vspace{2pt}
\noindent{\textbf{Training Configurations.}} We trained our MambaMOT and MambaMOT\textsuperscript{+} using Adam optimizer with $0.0001$ initial learning rate for 500 epochs with a batch size of $32$ on all of our experiments. We implemented the motion models using $2$ mamba blocks, with a hidden dimension of $64$, and expansion factor of $2$. For tracklet merging in MambaMOT \textsuperscript{+}, a temporal distance threshold of 50 frames and a spatial distance threshold of 50 pixels are employed to avoid merging unreasonable tracklet pairs. All the experiments are conducted on a single NVIDIA RTX 4080 GPU. 

\vspace{2pt}
\noindent{\textbf{Evaluation Metrics.}} We utilize commonly used tracking metrics, including HOTA~\cite{luiten2021hota} for its comprehensive evaluation of both detection and association accuracy; and CLEAR metrics~\cite{CLEARMOT}, where IDF1 and MOTA serve as the standard benchmark for tracking performance across various scenarios.

\subsection{Benchmark Evaluation}

\noindent{\textbf{DanceTrack.}} We compared MambaMOT with several different state-of-the-art methods on the DanceTrack test set in Table.~\ref{table:dance}. MambaMOT outperforms ByteTrack~\cite{zhang2022bytetrack} by a large margin of $8.2\%$ in HOTA when using the same detections, some trajectories visualizations are shown in Fig.~\ref{fig:fig3}. Furthermore, MambaMOT\textsuperscript{+} also demonstrates the ability to boost the performance, improving HOTA by $0.6\%$ and IDF1 by $1.0\%$.

\vspace{2pt}

\noindent{\textbf{SportsMOT.}} On the SportsMOT dataset, MambaMOT achieves the highest HOTA performance among all the learning-based methods and outperforms ByteTrack by \textbf{8.2\%} HOTA and \textbf{7.3\%} AssA under the same association method and detector setting, as shown in Table.~\ref{table:sportsmot}.

\subsection{Inference Speed}
Most of the previous learning-based methods such as MOTR\cite{zeng2022motr} suffer from a common drawback: their significant computational overhead and slower inference speeds, typically under 10 FPS. In contrast, MambaMOT addresses this limitation by integrating a lightweight motion model. As a result, it achieves a real-time inference speed of 28.8 FPS on a single GPU, demonstrating its efficiency and potential for real-world applications.

\begin{table}[t]

\caption{\textbf{Performance on DanceTrack test set.} We compared our performance with different Kalman filter-based and learning-based trackers on the DanceTrack dataset.}

\vspace{-12pt}

  \begin{center}
    {\small{
\begin{tabular}{lccccc}
\toprule

 Methods & HOTA & DetA & AssA & IDF1 & MOTA \\

\midrule
\textit{Kalman filter-based} &   &   &   &   &   \\
FairMOT \cite{zhang2021fairmot} & 39.7 & 66.7  & 23.8  & 40.8  & 82.2  \\
SORT \cite{bewley2016simple} & 47.9 & 72.0 & 31.2 & 50.8 & \textbf{91.8} \\
DeepSORT \cite{wojke2017simple} & 45.6 & 71.0 & 29.7 & 47.9 & 87.8 \\
ByteTrack \cite{zhang2022bytetrack} & 47.3 & 71.6 & 31.4 & 52.5 & 89.5 \\
OC-SORT \cite{cao2023observation} & 54.6 & 80.4 &  \textbf{40.2} & 54.6 & 89.6 \\

\midrule
\textit{Learning-based} &   &   &   &   &   \\
CenterTrack \cite{zhou2020tracking} & 41.8 & 78.1 & 22.6 & 35.7 & 86.8 \\ 
TraDes \cite{trades} & 43.3 & 74.5 & 25.4 & 41.2 & 86.2 \\
QDTrack \cite{fischer2023qdtrack} & 45.7 & 72.1 & 29.2 & 44.8 & 83.0 \\  
TransTrack \cite{sun2020transtrack} & 45.5 & 75.9 & 27.5 & 45.2 & 88.4 \\
MOTR \cite{zeng2022motr} & 54.2 & 73.5 & 40.2 & 51.5 & 79.7 \\
GTR \cite{zhou2022global} & 48.0 & 72.5 & 31.9 & 50.3 & 84.7 \\
MotionTrack \cite{xiao2023motiontrack} & 52.9 & \textbf{80.9} & 34.7 & 53.8 & 91.3 \\

\rowcolor{aliceblue}
MambaMOT & 55.5 & 80.8 & 38.3 & 53.9 & 90.1 \\

\rowcolor{aliceblue}
MambaMOT\textsuperscript{+} & \textbf{56.1} & 80.8 & 39.0 & \textbf{54.9} & 90.3 \\

\bottomrule
\end{tabular}
}}
\end{center}

\label{table:dance}

\vspace{-12pt}
\end{table}

\begin{table}[!t]

\caption{\textbf{Performance on SportsMOT test set.} Star symbol $^{*}$ denotes using the same detections generated by YOLOX. All the methods were trained using the SportsMOT training set only.
}
\vspace{-12pt}
  \begin{center}
    {\small{
\begin{tabular}{lccccc}
\toprule

 Methods & HOTA & DetA & AssA & IDF1 & MOTA \\

\midrule
\textit{Kalman filter-based} \\
FairMOT \cite{zhang2021fairmot} & 49.3 & 70.2 & 48.0 & 62.3 & 86.4 \\
ByteTrack$^{*}$ \cite{zhang2022bytetrack} & 62.0 & 77.0 & 49.9 & 68.7 & 93.9\\
OC-SORT$^{*}$ \cite{cao2023observation} & 70.2 & 85.9 & 57.4 & 70.4 & 93.9 \\
\midrule
\textit{Learning-based} \\
GTR \cite{zhou2022global} & 54.5 & 64.8 & 45.9 & 55.8 & 67.9 \\
CenterTrack \cite{zhou2020tracking} & 62.7 & 82.1 & 48.9 & 60.0 & 90.8\\
QDTrack \cite{fischer2023qdtrack}  & 60.4 & 77.5 & 47.2 & 62.3 & 90.1 \\
TransTrack \cite{sun2020transtrack} & 68.9 & 82.7 & 57.5 & \textbf{71.5} & 92.6 \\

\rowcolor{aliceblue}
MambaMOT$^{*}$ & 70.4 & 86.7 & 57.2 & 69.5 & 94.7 \\

\rowcolor{aliceblue}
MambaMOT\textsuperscript{+}$^{*}$ & \textbf{71.3} & \textbf{86.7} & \textbf{58.6} & 71.1 & \textbf{94.9} \\

\bottomrule
\end{tabular}
}}
\end{center}
\vspace{-18pt}
\label{table:sportsmot}

\end{table}
\section{Conclusions}
In this paper, we proposed MambaMOT, which address the limitations of traditional Kalman filter-based multi-object tracking by using learning-based motion models, showing significant performance improvements in various tracking evaluation metrics. The proposed method leverages state-space models to capture complex motion patterns, achieving comparable performance with the state-of-the-art methods on multiple public benchmarks.

% \input{tex/8_rebuttal}

% \newpage
\bibliographystyle{splncs04}
\bibliography{ref}

\end{document}